%% file: ICIP_main.tex
\documentclass{article}
\usepackage{spconf,amsmath,graphicx}
\usepackage{float}
\usepackage{hyperref}
\usepackage{booktabs} 

\input{section/macro}

\title{LEARNING UNCERTAINTY FOR SAFETY-ORIENTED SEMANTIC SEGMENTATION IN AUTONOMOUS DRIVING}
\name{Victor Besnier$^{1,2,3}$, David Picard$^3$, Alexandre Briot$^1$}
\address{1. Valeo, Créteil, France\\
2. ETIS UMR8051, CY Université, ENSEA, CNRS, Cergy France \\
3. LIGM, Ecole des Ponts, Univ Gustave Eiffel, CNRS, Marne-la-Vallée, France \\}

\begin{document}
\ninept
\maketitle
\input{section/abstract}
\input{section/introduction}
\input{section/related_work}
\input{section/method}

\begin{figure*}
    \centering
    \includegraphics[height=4.5cm,width=0.9\textwidth]{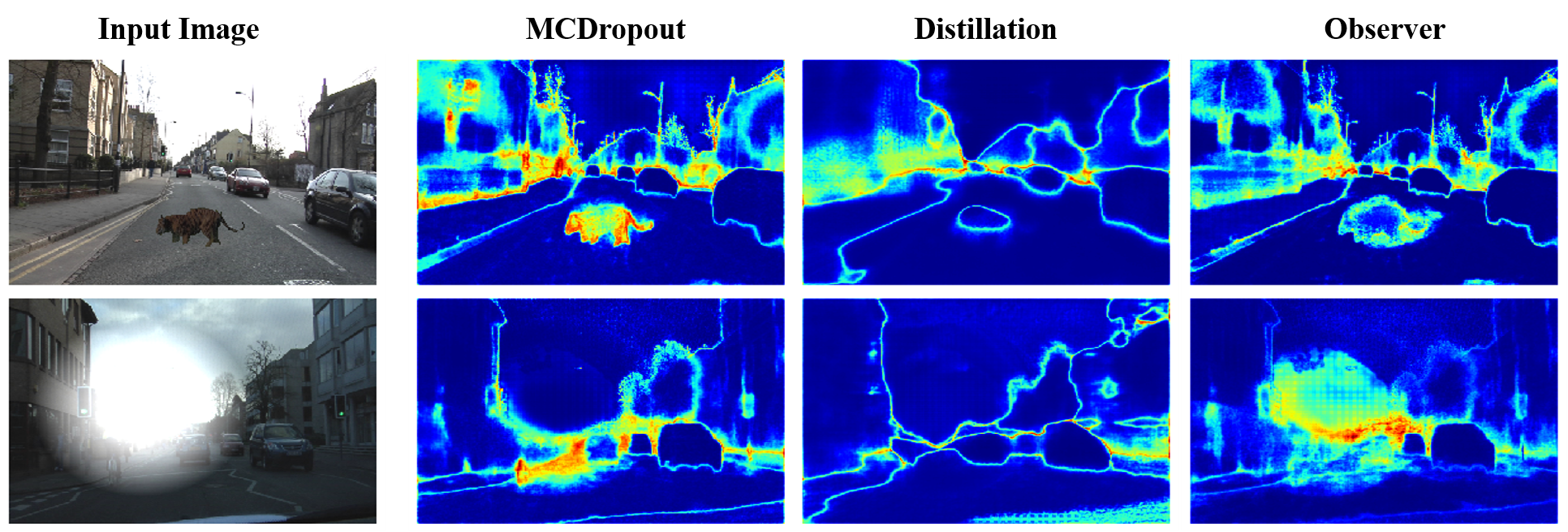}
    \caption{Uncertainty map. First line is with an OOD animal and in the image and second one is with glare. The observer is not only close to \mcdropout for epistemic detection, but is also capable of detect aleatoric uncertainty, while distillation is perform badly.}
    \label{fig:table_image}
    \vspace{-4mm}
\end{figure*}
\input{section/results}
\input{section/conclusion}

\newpage
\bibliographystyle{IEEEbib}
\bibliography{citation}
\end{document}

%% file: section/macro.tex
\usepackage{amsmath}
\usepackage{amssymb}
\usepackage{dsfont}
\usepackage{bm}
\usepackage{xcolor,soul,colortbl}
\usepackage{array}
\usepackage{multirow}
\usepackage{subcaption}
\usepackage{dashrule}
\usepackage{graphicx}
\usepackage{breqn}
\usepackage{enumitem}
\usepackage{xspace}
\usepackage{float}

\newcommand{\mcdropout}{MC Dropout\xspace}
\newcommand{\obsnet}{ObsNet\xspace}
\newcommand{\eg}{\emph{e.g.},\xspace} 
\newcommand{\ie}{\emph{i.e.},\xspace} 
\newcommand{\etc}{\emph{etc.}\xspace}

%% file: section/abstract.tex
\begin{abstract}
In this paper, we show how uncertainty estimation can be leveraged to enable safety critical image segmentation in autonomous driving, by triggering a fallback behavior if a target accuracy cannot be guaranteed.
We introduce a new uncertainty measure based on disagreeing predictions as measured by a dissimilarity function. We propose to estimate this dissimilarity by training a deep neural architecture in parallel to the task-specific network. It allows this observer to be dedicated to the uncertainty estimation, and let the task-specific network make predictions.
We propose to use self-supervision to train the observer, which implies that our method does not require additional training data.
We show experimentally that our proposed approach is much less computationally intensive at inference time than competing methods (\eg \mcdropout), while delivering better results on safety-oriented evaluation metrics on the CamVid dataset, especially in the case of glare artifacts.
\end{abstract}
\begin{keywords}
Uncertainty, Segmentation, Autonomous Driving
\end{keywords}



%% file: section/introduction.tex
\section{Introduction}\label{section:Introduction}
With the recent development of deep learning, neural networks have proven to reach or even beat human level performance at solving complex visual tasks necessary for autonomous driving such as object detection and image recognition \cite{xie_2017_aggregated,chen2017rethinking, He_2017_ICCV}. However, when it comes to safety critical applications, the use of neural networks raises huge challenges still to be unlocked. Building models able to outperform humans in dealing with unsafe situations or in detecting operating conditions in which the system is not designed to function remains an open field of research. For example, although \emph{softmax} outputs class conditional probabilities, it produces misleading high confidence outputs even in unclear situations~\cite{mozejko2018inhibited}. 

\begin{figure}[ht]
\centering
\begin{subfigure}{.23\textwidth}
  \centering
  \includegraphics[height=2.5cm, width=.95\linewidth]{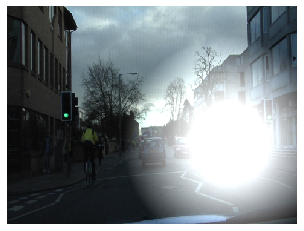}
  \caption{Input Image}
  \label{fig:sfig.95}
\end{subfigure}%
\begin{subfigure}{.23\textwidth}
  \centering
  \includegraphics[height=2.5cm, width=.95\linewidth]{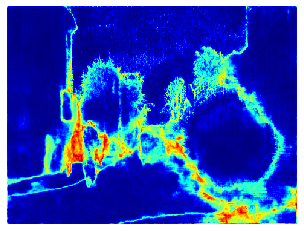}
  \caption{\mcdropout}
  \label{fig:sfig4}
\end{subfigure}%
\\
\begin{subfigure}{.23\textwidth}
  \centering
  \includegraphics[height=2.5cm, width=.95\linewidth]{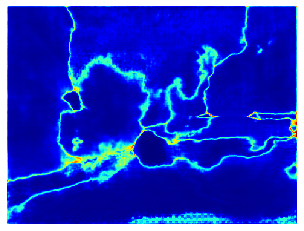}
  \caption{Distillation}
  \label{fig:sfig5}
\end{subfigure}%
\begin{subfigure}{.23\textwidth}
  \centering
  \includegraphics[height=2.5cm, width=.95\linewidth]{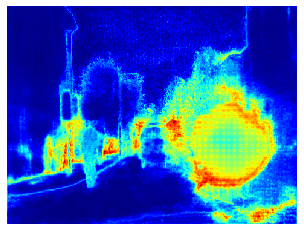}
  \caption{Ours}
  \label{fig:sfig6}
\end{subfigure}
\caption{Uncertainty maps for a noisy image. \textbf{(a)} Image with artificial sun glare. \textbf{(b)} can detect epistemic uncertainty but not aleatoric uncertainty caused by the glare. \textbf{(c)} is fast in inference but do not capture aleatoric uncertainty. \textbf{(d)} Our proposed observer detects epistemic and aleatoric uncertainty and is as fast as distillation.}
\label{fig:teaser}
\vspace{-4mm}
\end{figure}

This is all the more dramatic in autonomous driving \cite{Michelmore2018EvaluatingUQ} applications where the confidence of the prediction is safety critical. Demonstration of safety for AI components is a key challenge addressed by SOTIF \cite{SOTIF} which focuses on external events (external to the system) that are not correctly handled by the system (e.g. weather conditions, user driving tasks, road users, ...). In such systems, detecting uncertain predictions is a key trigger to produce a safe behavior by interrupting the current process and starting a fallback process (\textit{e.g.}, human intervention) instead of risking a wrong behavior.

In this paper, we focus on predicting the uncertainty of a given semantic segmentation network in an autonomous driving context. 
Our main contribution is a safety oriented uncertainty estimation framework that consists of a deep neural network observer (as proposed in \cite{SAFAD}) running in parallel to the target segmentation network.
This auxiliary network is trained using self-supervision to output predictions that are similar to the target network when the target network is certain, and completely different outputs when the prediction is uncertain.
The uncertainty is then measured as the dissimilarity between the target network output and the observer output and we empirically show it produces a good proxy for measuring the uncertainty to detect safety critical predictions.
Thus, our framework has the following properties that highlight its relevance:
\begin{itemize}[noitemsep,topsep=0pt]
    \item It improves over other uncertainty estimation methods at detecting safety critical predictions;
    \item It is trained in a simple self-supervised fashion, meaning that no expensive annotations are required;
    \item It does not require retraining the target network and can work with any off-the-shelf network;
    \item It is fast, the processing of the observer happens in parallel to the target network, meaning it has a reduced overhead compared to other uncertainty estimation methods.
    \item It is able to detect uncertain predictions arising from glare artifacts.
\end{itemize}


%% file: section/related_work.tex
\section{Related work}\label{section:related_work}
Semantic segmentation is an important task in autonomous driving \cite{Cordts2016Cityscapes}. However, wrong predictions can lead to disastrous events. As such a confidence score has to be computed to detect unreliable predictions at the pixel level. In that context, uncertainty estimation is safety critical vision task \cite{ijcai2017-661}.

It is well known that the \emph{softmax} output is an overconfident score that does not correctly estimate uncertainty \cite{guo_2017}. Even techniques to calibrate the output score like temperature scaling \cite{guo_2017} show limitation to detect out-of-distribution sample. On the contrary, methods estimating uncertainty as the variations among multiple predictions such as \cite{Lakshminarayanan_2017} or \cite{Gal2016Dropout} have shown promising results \cite{snoek2019can}. 


\textbf{Uncertainty from Bayesian Inference}. In \cite{Gal2016Dropout}, the authors propose to use Bayesian Inference to estimate the epistemic uncertainty of a model. They use dropout during training to approximate Bayesian Inference in deep Gaussian processes. During inference, they perform multiple forward passes with dropout to obtain multiple predictions. They propose to use the entropy of the average prediction as the measure of epistemic uncertainty, and show it produces promising results. However, the computational cost is very high due to the multiple forwards and it cannot model aleatoric uncertainty. \cite{teye2018bayesian}, \cite{KendallGal_2017} and \cite{zhang2017noisy} propose similar stochastic techniques to estimate epistemic uncertainty, with the same drawbacks.

\textbf{Dropout Distillation}. In order to avoid the computational cost of the multiple forwards, \cite{Gurau_2018} and \cite{malinin_ensemble_2019} proposes to use distillation. Distillation considers two networks, a generally large one called Teacher and a generally smaller one called Student. The Student is trained so as to mimic the Teacher. In the case of uncertainty estimation, the Student is trained to capture the variance produced by the dropout in the Teacher \cite{Gurau_2018}. Although distillation removes the burden of the multiple passes, it is a much more complex problem since a single network has to solve two tasks with a single head. 

\textbf{Sample Free Epistemic Uncertainty Estimation}. In \cite{postels2019sampling}, the authors propose a new framework for epistemic uncertainty without multiple forward passes by measuring the propagation of the variance caused by noise in the network. To achieve this efficiently, they propose a simplification of the propagation effect in a convolution layer followed by ReLU that limits memory requirement. \cite{choi_uncertainty-aware_2018}, \cite{le_uncertainty_2018} proposes similar ideas based on density estimation networks. However, similarly to all other methods, these sample-free techniques only model epistemic uncertainty and do not consider aleatoric uncertainty.

Contrarily to the related work, our proposed framework is able to capture both epistemic and aleatoric uncertainty. It is most closely related to distillation based methods in that we also use a second network. However, it is dedicated to uncertainty estimation, which we show is crucial. It is not trained to capture the variance of the main network, but is instead trained to output a second prediction that can be compared to the main prediction to estimate uncertainty. We experimentally show that it leads to better results. 

%% file: section/method.tex
\section{Proposed Method}\label{section:method}
\begin{figure}
    \centering
    \includegraphics[width=\linewidth]{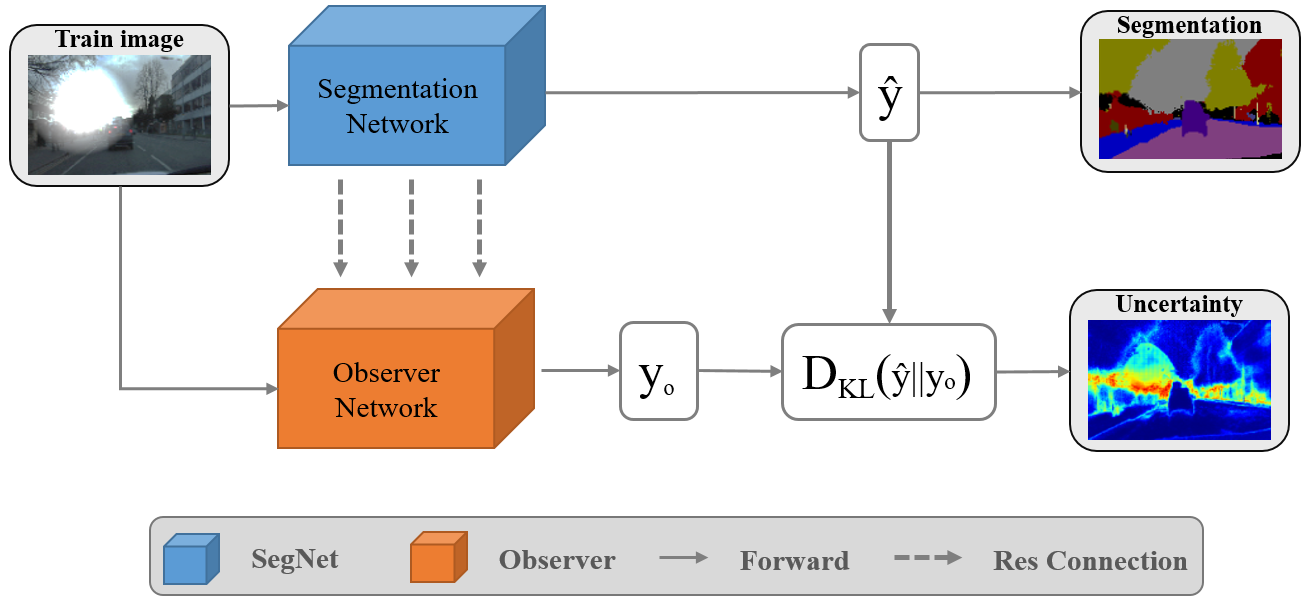}
    \caption{Method architecture. $\hat{y}$ main prediction, $y_o$ observer output. Our architecture is dedicated to uncertainty measurement.}
    \label{fig:archi}
    \vspace{-4mm}
\end{figure}
For safety purposes, it is often better to not make any decision rather than making a decision that cannot be guaranteed. Given a trigger that detects decisions that cannot be guaranteed, a safe system could then stop its current process and start a fallback process (\eg human intervention). Such a trigger divides the predictions $\hat{y}$ of a given neural network into 2 classes: the class of \emph{certain} predictions associated with the class $c=+1$, for which an average error rate can be guaranteed; and the \emph{uncertain} class, associate with the label $c=-1$, for which no average guarantee can be obtained. More formally, we have the following property:
\begin{equation}\label{eq:1}
    \mathbb{E}_{\hat{y}|c=+1}[l(\hat{y}, y)] \leqslant \epsilon,
\end{equation}
with $\hat{y}|c=+1$ the distribution of \emph{certain} predictions, y the ground truth and $l(.,.)$ the loss function of the target application. 

We propose to use uncertainty estimation to obtain such safety trigger. Given a function $u(\hat{y})$ that estimates the uncertainty of a prediction $\hat{y}$, we define a safety threshold $\delta$ such that predictions under the threshold are in the \emph{certain} set: 
\begin{equation}\label{eq:2}
    u(\hat{y}) \leq \delta 	\Rightarrow c=+1
\end{equation}
In practice, given an error rate threshold $\epsilon$ and its corresponding uncertainty threshold $\delta$, we evaluate uncertainty functions by the recall, that is, the proportion of samples that are deemed certain.
\begin{figure*}[ht]
    \centering
    \includegraphics[height=4.5cm, width=0.96\textwidth]{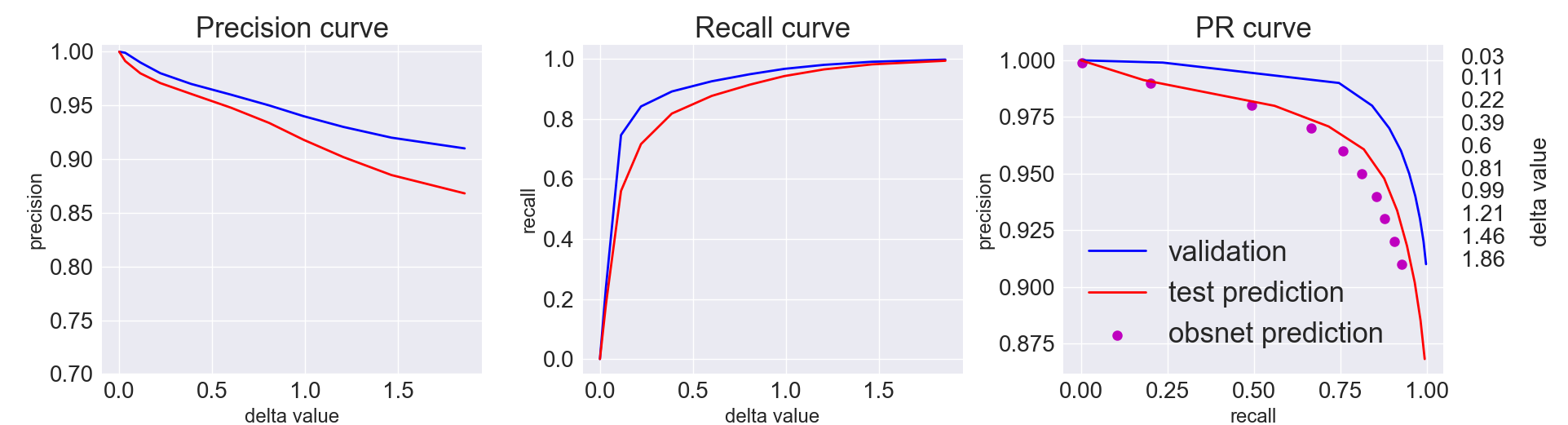}
    \caption{Precision and Recall depending on the uncertainty threshold $\delta$ for epistemic uncertainty. Tune the hyper-parameter $\delta$, allows to select which examples are in the \emph{certain} prediction set (\ie $c=+1$).}
    \label{fig:ablation_delta}
    \vspace{-4mm}
\end{figure*}

\subsection{Dissimilarity based uncertainty} 
To obtain $u(.)$, we propose to model the uncertainty as the dissimilarity between several predictions. Let $\hat{y}$ be the prediction of the neural network we are analyzing, and let $y_a$ be an additional prediction without any assumption on how it is obtained (\eg additional forward of a stochastic model, ensemble, oracle, \etc).
We propose that measuring the dissimilarity between $\hat{y}$ and $y_a$ gives us a sense of the uncertainty regarding prediction $\hat{y}$.
In the case of a regression problem, the dissimilarity can be measured as the distance between the two predictions.
In this paper, we focus on softmax classification which is used in semantic segmentation and is  the well studied in uncertainty estimation. It that case, $\hat{y}$ and $y_a$ being the probability distribution over the different possible classes (softmax outputs), we propose to use the KL divergence to define the uncertainty function $u(.)$:
\begin{equation}\label{eq:3}
    u(\hat{y}) = D_{KL}(\hat{y}||y_a).
\end{equation}
In other words, if $\hat{y}$ and $y_a$ produce similar outputs, the uncertainty of $\hat{y}$ is low, whereas if $\hat{y}$ and $y_a$ produce dissimilar (or disagreeing) outputs, the uncertainty of $\hat{y}$ is high.

\subsection{Learning to predict \texorpdfstring{$y_a$}{ya}}
Unfortunately, it is clear that the additional prediction $y_a$ may not be available at inference time. Our main contribution solves this problem by introducing a second predictor, the observer with output $y_o$ which we use in place of $y_a$. Training $y_o$ to precisely regress the predictions $y_a$ can be a difficult learning problem. Moreover, in case of \emph{uncertain} predictions, it is not required that $y_o$ perfectly matches $y_a$. Instead, $y_o$ only has to be sufficiently different from $\hat{y}$ to produce a large KL divergence, just as $y_a$ would have done. Therefore, we propose to train $y_o$ using a self-supervised classification problem distinguishing between \emph{certain} and \emph{uncertain} predictions.

In practice, given a training set of pairs $(\hat{y}, y_a)$, we use the safety threshold $\delta$ to split the pairs into \emph{certain} $c=+1$ and \emph{uncertain} $c=-1$ classes, and we train our predictor \obsnet to minimize $D_{KL}(\hat{y}||y_o)$ for $c=+1$ and maximize $D_{KL}(\hat{y}||y_o)$ for $c=-1$ by optimizing the following problem:
\begin{equation}\label{eq:4}
    \min_{\theta}D_{KL}(\hat{y}||y_o)^c,
\end{equation}
with $\theta$ the parameters of the observer. 
We argue that this objective is much easier to optimize than regressing $y_a$ since the output $y_o$ has many more degrees of freedom in the \emph{uncertain} case. Instead of being force to predict the exact same class as $y_a$, $y_o$ can predict any class different from the one predicted by $\hat{y}$.

Usually, uncertainty is classified into two different classes: Epistemic and Aleatoric which we both propose to estimate using different additional predictions $y_a$. 

\subsection{Epistemic Uncertainty}
Epistemic uncertainty is associated with the model uncertainty. It capture the lack of knowledge about the process that generated the data.
To estimate epistemic uncertainty, we propose to use \mcdropout as the additional prediction $y_a$. We compute the additional prediction $y_a$ as the average of T forward passes with dropout. Note that our training setup is entirely self-supervised as the labels $c$ are obtained using $D_{KL}$ over forward passes of the target network only.

We show on \autoref{fig:ablation_delta} what setting a specific threshold $\delta$ implies in terms of precision and recall for \emph{certain} ($c=1$) predictions. As we can see, the observer is perfectly able to recover the specific operating points of the uncertainty obtained by the original \mcdropout additional prediction.

\subsection{Aleatoric uncertainty}
The aleatoric uncertainty is associated with the natural randomness of the input signal~\cite{KendallGal_2017}. More precisely, in this work, we focus on heteroscedastic uncertainty, which is the lack of visual features in the input data (e.g. sun glare, occlusion, ...).
We propose to artificially create such cases by adding random glare noise to the input image. The prediction $\hat{y}$ is obtained by a single forward pass on the noisy image, while the additional prediction $y_a$ is obtained by a single forward pass on the clean image.

The uncertainty function $u(\hat{y})=D_{KL}(\hat{y}||y_a)$ then establishes whether the noise we added led to aleatoric uncertainty or not. Thanks to $\delta$, we label the pair with either $c=+1$ or $c=-1$. As for epistemic uncertainty, we train our auxiliary network on the labeled pairs $(\hat{y}, y_a)$ using \autoref{eq:4}.


\subsection{Observer Architecture}
To implement the uncertainty output $y_o$, we propose to add an observer network in parallel to the main network as shown on \autoref{fig:archi}. The additional network mimics the architecture of the main network. It takes the image as input as well as the activation maps of the main network as additional inputs thanks to  residual connections. The uncertainty is obtained by computing the KL divergence with the main network output (\ie $D_{KL}(\hat{y}||y_o)$).

%% file: section/results.tex
\section{Results}\label{section:results}
\subsection{Datasets, Metrics and Baselines}
In this section, we evaluate our method on CamVid \cite{BrostowFC:PRL2008}, a dataset for road image segmentation. To compare the results, we adopt safety oriented metrics. We measure the recall when the precision is equal to 95\% (e.g. R@P=0.95). Better uncertainty measures ought to achieve higher recall. Moreover, we propose a "safety trigger rate metrics" which is the percentage of images in the dataset with a coverage of certain prediction over a threshold. For instance "Trigger 75\%" is the percentage of images in the test set where the coverage of safe prediction is above 75\%. And we also report the area under the curve (AuPR) which is a threshold independent metric. We compare several methods:
\begin{itemize}[noitemsep,topsep=0pt]
    \item \textbf{Softmax \cite{hendrycks_benchmark_2019}}: One minus the maximum of the prediction.
    \item \textbf{Void Class (VC)\cite{blum2019fishyscapes}}: Void/unknown class prediction for segmentation.
    \item \textbf{\mcdropout \cite{Gal2016}}: We consider this as baseline. We use $T=50$ and $T=2$ forward passes.
    \item \textbf{MCDA\cite{ayhan_test-time_2018}}: Data augmentation such as geometric and color transformations added during inference time to capture aleatoric uncertainty.
    \item \textbf{Distillation \cite{Gurau_2018}}: We propose two variants of the distillation: supervised (\ie using Teacher outputs and cross-entropy with ground truth for training) and unsupervised (\ie the student only regress the teacher output).
    \item \textbf{Observer}: Our proposed method, with KL divergence based uncertainty training. We use two oracle: \mcdropout (self-sup) and Ground Truth (GT). The Ground Truth variant uses ground truth labels as additional prediction $y_a$ and is used to measure the influence of having a self supervised setup.
\end{itemize}

For all our segmentation experiments we use a Bayesian SegNet \cite{badrinarayanan2015segnet}, \cite{kendall2015bayesian} with dropout as the main network. Therefore, our \obsnet follows the same architecture as this SegNet. 

For each mini-batch, we compute 50 forward passes with dropout to compute $y_a$ for epistemic uncertainty. For aleatoric uncertainty, we add noise to every input image to obtain $\hat{y}$ and use the noiseless image to calculate $y_a$. The final loss is the sum of the aleatoric uncertainty loss and epistemic loss. The observer is trained with SGD with momentum and weight decay, by minimizing the loss \autoref{eq:4}. It is trained for 50 epochs and we keep the best performing network on the validation set. 

To train distillation for aleatoric uncertainty estimation, we change the training set up to be fair with our method. When the student gets an noisy image, it is train to output the same prediction as the teacher given a de-noised image. 

\subsection{Epistemic uncertainty} 
We first report results on \autoref{tab:classic_table}. Our method with GT performs best or similar to \mcdropout in all the safety metrics. Our self-supervised variant method is better than \mcdropout at comparable computational cost\footnote{Our framework is equivalent to \mcdropout T2 and 20 times faster than T50 on a GeForce RTX 2080 Ti.}. Distillation alone offers low performances, which is due to the complexity of regressing the exact \mcdropout outputs and difficulty to capture uncertainty and class prediction.
\begin{table}
 \centering
  \begin{tabular}{lccc} \toprule
	 Method & R@P=0.95 & AuPR & Trigger 75\%\\ \hline
     MCDropout T50 & 88.2 & \textbf{97.9} & \textbf{81.5}\\
	 MCDropout T2 & 85.9 & 97.5 & 76.4\\ 
	 Softmax & 81.0 & 96.5 & 67.4  \\ 
	 Void Class & 64.1 & 95.2 & 39.9 \\ \hline
	 Distill supervised & 65.9 & 95.3 & 31.8 \\ 
	 Distill self-sup & 68.6 & 96.0 & 34.3 \\ \hline
	 Observer self-sup \textbf{(ours)} & 87.3 & 97.7 & 76.8 \\ 
	 Observer GT \textbf{(ours)} & \textbf{89.3} & \textbf{97.9} & \textbf{81.5} \\ \bottomrule
  \end{tabular}
  \caption{Evaluation of epistemic uncertainty, best method in bold.}
  \label{tab:classic_table}
  \vspace{-5mm}
\end{table}

\subsection{Aleatoric uncertainty}
We added glare in the image: a very important increase of brightness in an ellipse of random size and coordinates on the image. As we can see on \autoref{tab:aleatoric_table}, \mcdropout suffers dramatic failures and is unable to obtain high recall for the glare. This is to be expected since \mcdropout is not designed to capture aleatoric uncertainty. MCDA performs the best on glare among set-up without additional network. As with epistemic uncertainty, distillation does not succeed in producing good results. In contrast, our observer obtains significantly better results in all the considered cases. Overall, ours framework significantly outperforms \mcdropout and data-augmentation based method for aleatoric uncertainty while being much less computationally expensive.

To evaluate the generalization capabilities of our method, we train on patch noise (\ie random patch on the images are added during training) and evaluate on glare (bottom of \autoref{tab:aleatoric_table}). As we can see, although training on a different noise decreases the performance compared to the full training, we still outperforms \mcdropout, distillation and MCDA by a large margin. This shows the capacity of the observer to generalize the unseen noises.

\begin{table}
 \centering
  \begin{tabular}{lccc@{\hskip2.3mm}c@{\hskip2.3mm}c@{\hskip2.3mm}} \toprule
    Method                             & Train & Test     & R@P   & AuPR & Trigger \\ \hline
	\mcdropout T50                     & -     & glare    &  0.1 & 83.9 & 18.9  \\
	\mcdropout T2                      & -     & glare    &  0.0 & 83.7 & 15.5  \\ 
	Softmax                            & -     & glare    &  1.0 & 90.5 & 17.2  \\ 
	Void Class                         & -     & glare    &  0.3 & 82.7 &  8.2  \\
	MCDA T50                           & -     & glare    & 44.7 & 91.7 & 14.3  \\ \hline
	Distill supervised                 & glare & glare    &  0.0 & 83.3 &  2.2  \\ 
	Distill supervised                 & patch & glare    &  0.3 & 85.7 &  4.1  \\ 
	Distill self-sup                   & glare & glare    &  1.6 & 84.1 &  3.0  \\
    Distill self-sup                   & patch & glare    &  1.9 & 85.5 &  4.3  \\ \hline
	Obs self-sup \textbf{(ours)}       & glare & glare    & 68.4 & 95.3 & 23.2  \\
	Obs GT \textbf{(ours)}             & glare & glare    & \textbf{79.4} & \textbf{96.6} & \textbf{39.2}  \\
	Obs self-sup \textbf{(ours)}       & patch & glare    & 46.7 & 92.3 & 15.1  \\
    Obs GT \textbf{(ours)}             & patch & glare    & 54.5 & 93.8 & 26.2  \\ \bottomrule

  \end{tabular}
  \caption{Evaluation of aleatoric uncertainty.}
  \label{tab:aleatoric_table}
  \vspace{-5mm}
\end{table}

We show qualitative uncertainty maps of \autoref{fig:table_image}. \mcdropout and ours output very similar maps for epistemic uncertainty, while outperform others methods on noisy images.

%% file: section/conclusion.tex
\section{Conclusion}\label{section:Conclusion}
In this paper, we have present a simple and effective method to estimate uncertainty. We introduce a safety oriented context where the uncertainty is used to trigger a safety signal when a given error rate cannot be met. Contrarily to ensemble methods, our framework requires a single forward pass making it computationally efficient. Contrarily to distillation based methods,our observer relies on self-supervised classes and is much more effective. With experiments on CamVid, we show that our method obtains improved results compared to competing approaches.